\documentclass[a4paper,10pt,3p,authoryear]{elsarticle}

\usepackage{amsmath,amssymb,amsfonts}
\usepackage{algorithmic}
\usepackage{graphicx}
\usepackage{textcomp}
\usepackage{xcolor}

\usepackage{multirow}
\usepackage{makecell}
\usepackage{booktabs}
\usepackage{url}
\usepackage{subcaption}
\usepackage{lineno}
\usepackage{hyperref}

\usepackage{framed}
\usepackage{nomencl} 
\usepackage[T1]{fontenc}  % Some author names with different letters
\usepackage{setspace}

\hypersetup{hidelinks}

\journal{Computers and Electronics in Agriculture}

\begin{document}

% TODO: remove this for submission
% Only for ArXiv
\makeatletter
\def\ps@pprintTitle{%
  \let\@oddhead\@empty
  \let\@evenhead\@empty
  \let\@oddfoot\@empty
  \let\@evenfoot\@oddfoot
}
\makeatother

\begin{frontmatter}

\title{A drone that learns to efficiently find non-uniformly distributed objects in agricultural fields: from simulation to the real world}

\author[1]{Rick van Essen}
\author[1]{Eldert van Henten}
\author[1]{Gert Kootstra\corref{cor1}}
\ead{gert.kootstra@wur.nl}

\affiliation[1]{
    organization={Agricultural Biosystems Engineering, Department of Plant Sciences, Wageningen University and Research, 6700 AA},
    city={Wageningen},
    country={The Netherlands}
}
\cortext[cor1]{Corresponding author.}

\begin{abstract}  % Max 250 words
Drones are promising for data collection in precision agriculture but are limited by battery capacity. Drone paths are usually planned using full coverage planners, even though this is not always required. This paper presents a drone path planner trained with Reinforcement Learning (RL) to detect as many objects as possible with a minimal flight path length. The agent uses low-quality prior knowledge derived from a high-altitude full coverage flight as guidance. The agent was trained in simulation, modeling object distributions, drone movement, field geometry, detection errors, and uncertain prior knowledge. Combined with a flight controller and object-detection network, it controls flight direction, terminates flights, and can be deployed on a real drone. It was evaluated across six levels of realism, from pure simulation to real-world drone flights, to quantify the simulation-to-reality gap. The agent achieved a 57\% shorter flight path than a full coverage planner in simulation (13\% lower recall) and a 38\% shorter flight path on real-world orthomosaic data (21\% lower recall). In real-world drone flights, the agent found 73\% and 23\% of the objects in trials~1 and~2, respectively; the lower real-world performance was mainly attributed to prior knowledge quality. Although framed as a weed-detection task, the approach is expected to generalize to other agricultural tasks with non-uniformly distributed objects and tolerance to false negatives, though further research is needed before practical deployment.
\end{abstract}

% \begin{highlights} % Max 85 characters each, 3-5 items
%     \item We present the real-world deployment of a simulation-trained RL path planner
%     \item The simulation-to-reality gap is quantified across six levels of realism
%     \item In simulation, the agent achieves a 57\% shorter path at 13\% lower recall
%     \item On real-world data, the agent achieves a 38\% shorter path at 21\% lower recall
%     \item Prior knowledge quality is the dominant factor in real-world performance
% \end{highlights}


\end{frontmatter}

\section{Introduction}
\noindent
Drones are a promising tool in precision agriculture to gather information about crops \citep{Guebsi2024}. Data gathered by drones can, for example, be used for early disease detection \citep{Bouguettaya2023}, weed detection \citep{Gasparovic2020}, and soil quality assessment \citep{Nasi2023}. A limitation of using drones is their limited battery capacity \citep{Rejeb2022,Guebsi2024}, which limits the area that can be covered in a single flight. Most agricultural drones use full coverage planners to map the whole field with an equal spatial distribution. However, this is not needed for all applications. For example, some weed species and plant diseases are distributed in spatially distinct patches instead of uniform randomly \citep{Garibay2001, Campbell1985}. For such applications, it is not always required to find all objects, so a shorter flight path can be sufficient to find most objects. Using adaptive path planners that are reactive to the environment can make data collection more efficient and less time-consuming by exploiting these spatial patterns e.g., \citep{Popovic2024, vanEssen2025RLPlanning}. 

Adaptive path planners make decisions during flight to adapt the flight path when new information becomes available \citep{Popovic2024}. These decisions can be made rule-based or learning based. For example, \citet{vanEssen2025AdaptivePathPlanning} proposed a rule-based adaptive path planner for weed localization in agricultural fields, achieving a 37\% shorter flight path than full coverage with only a 2\% lower F1-score on real-world data. The advantages of learning based planning methods over rule-based methods are their ability to naturally account for uncertainties and changes in an environment \citep{Popovic2024}, without requiring explicit redesign for each new application. One approach for training a learning-based adaptive path planner is reinforcement learning (RL). RL can be used to train an agent to make sequential decisions in an environment by optimizing an action-value function that converts the state of an environment into an action that gives the highest long-term reward \citep{Sutton2018}. 

\begin{figure}[!t]
   \centering
   \includegraphics[width=0.5\columnwidth]{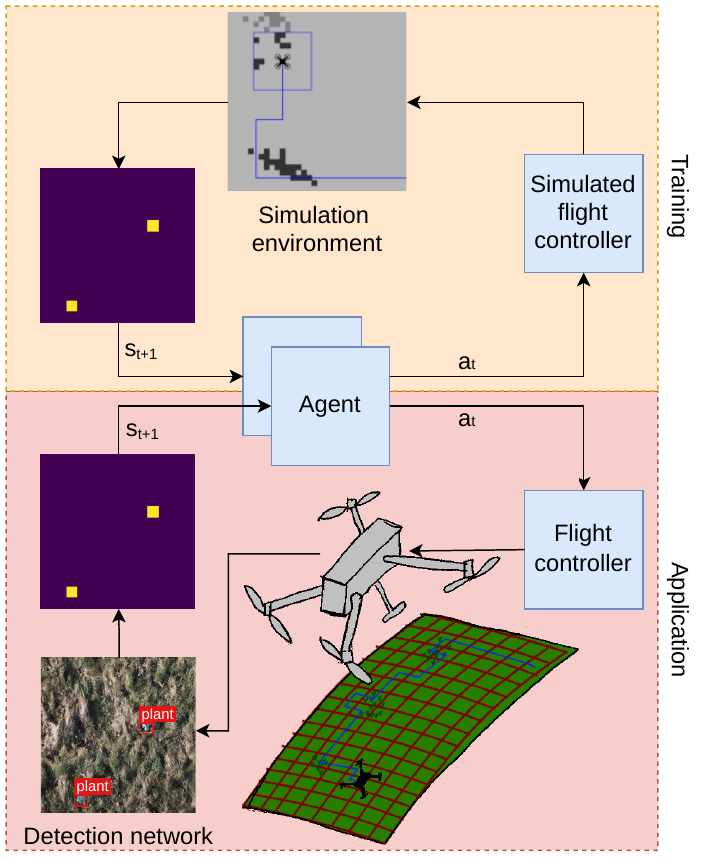}
   \caption{Overview of the learning-based adaptive path planner, which learns a policy in simulation that can be applied in the real world. The agent is trained in an abstract simulation using a deep Q-Network to select the best action $a_t$ given state $s_t$ that maximizes reward $r_t$ in the long term. During application, the drone's flight controller executes the action $a_t$ and the output from the detection network is converted to state $s_{t+1}$.}
   \label{fig:rl-drone}
\end{figure}

RL has gained more attention for path planning for drones over the last years \citep{Azar2021,Tu2023,vanEssen2025RLPlanning}. Most RL path planners assume full prior knowledge about the locations the drone has to visit or the obstacles to avoid; however, this prior knowledge is often uncertain or even absent. \Citet{Tu2023} proposed a path planner that uses RL to deviate from its flight path when detecting an obstacle. \Citet{vanEssen2025RLPlanning} introduced an RL-based agent, extending the work of \citep{Theile2021}, for path planning to localize objects of interest, such as weeds, in a field. The agent used uncertain prior knowledge about the locations of the weeds in combination with the output of a detection network. Such prior knowledge can be derived from various sources, such as high-altitude surveys, previous flights, or data from prior years, and does not need to be accurate. In a simulated environment, they showed that the RL agent outperformed a full coverage planner by localizing 80\% of the weeds in less than half of the flight time. They, however, did not test the applicability in the real world.

Training an RL agent using real-world data is infeasible, since the agent needs to explore a gigantic state-action space \citep{Zhao2020}. For this reason, training in simulation is a common approach for RL \citep{Gugan2023}. Figure~\ref{fig:rl-drone} shows the training and application workflow of the approach presented in this work. The RL agent is trained on an abstract simulation environment that simulates object distributions in the field, the movement of the drone over the field, and the output of an object detection network, including detection errors. During application, an object detection network is used to detect objects in images and convert them into the representation understood by the agent. The actions of the RL agent, such as moving forward, backward, left and right, can be translated to GPS coordinates and executed by the drone. In this way, the detection network captures most of the variance of the real world, and since its output is relatively simple to simulate, the complexity of the training environment is reduced compared to approaches that train on raw sensor data.

However, even with this reduced complexity, a gap between the simulated and real environment remains, and a sufficiently realistic simulator is still needed to transfer a simulation-trained RL policy to the real world. Different methods exist to decrease the gap between simulation and reality. Domain randomization is one of the most widely used techniques in RL, where instead of precisely parameterizing the world to match the real-world distribution of data, enough variability is included in the simulator to generalize to real-world data \citep{Zhao2020}. 

In this paper, weed detection is used as the use-case, where the goal is to efficiently localize weeds in an agricultural field using a drone. The objective of this paper is to extend the simulation-trained RL agent proposed in \citet{vanEssen2025RLPlanning} with a more diverse and randomized training environment to reduce the sim-to-real gap, to evaluate the resulting agent on real-world data, and demonstrate its use on a real drone. Specifically, this paper makes the following contributions: (1) we extend the original simulator with larger and randomly shaped fields and an expanded set of flight actions, (2) we apply domain randomization by training the network on a large variety of different object distributions, number of objects and field shapes to decrease the sim-to-real gap, (3) we incorporate prior knowledge derived from real-world data, (4) we evaluate the agent's performance on a dataset of 28 orthomosaics across six systematic levels of realism that progressively replace simulated components (object positions, prior knowledge, observations, and the flight controller) with their real-world counterparts, revealing each component's contribution to the sim-to-real gap, (5) we demonstrate real-world flights in which the RL agent directly controls the drone’s navigation decisions, and (6) we analyze the performance differences in detail between these flights and the orthomosaics. To support further research in this area, we make our code, data, and evaluation pipeline publicly available.

\section{Material and methods}
\noindent
In this section, we describe the problem formulation (Section~\ref{sec:mdp}), the network architecture and training (Section~\ref{sec:network_architecture}), the simulation environment (Section~\ref{sec:simulation_env}), the real-world application of the agent (Section~\ref{sec:application}), the orthomosaic datasets used for evaluation (Section~\ref{sec:datasets}), the six levels of realism used to quantify the sim-to-real gap (Section~\ref{sec:levels_of_realism}), the performance metrics (Section~\ref{sec:performance_metrics}), and the experiments (Section~\ref{sec:experiments}).

\subsection{Problem formulation}
\label{sec:mdp}
\noindent
The environment is modeled as a Markov Decision Process (MDP) \citep{Sutton2018}, containing a state-space, action-space and reward.

\subsubsection{State-space}
Inspired by \citet{Theile2021}, we model the state-space as a drone-centered global and local map, together with the drone's battery level. The global map gives low-resolution information of the whole field, whereas the local map contains high-resolution information about the camera's field of view. The global and local map share three layers: a field-border map indicating the borders of the field, a map containing the locations of the already found objects, and, in addition to \citet{vanEssen2025RLPlanning}, a coverage map indicating the grid cells that are already seen by the drone to discourage revisiting previously observed areas. Additionally, the global map contains a layer with the prior knowledge map, whereas the local map contains a layer for the observation map. The global map is padded to a size of $(2M - 1) \times (2M - 1)$ to make it drone-centered. The local map has a size of $N \times N$. The battery level is calculated by $b = b_\textrm{init} - n \cdot b_\textrm{step}$ where $n$ is the number of flight steps, $b_\textrm{init}$ the initial battery level, and  $b_\textrm{step}$ the battery usage for each step. An example of the global and local map is shown in Figure~\ref{fig:network_architecture} as input for the network.

\subsubsection{Action-space}
The action-space of the agent contains the following actions: fly north, fly south, fly west, fly east and land. In addition to previous work, we also added the four diagonal actions: fly north east, fly north west, fly south east, and fly south west, to allow more direct routing across the larger and randomly shaped fields used in this work. Each action moves the drone by $d_\textrm{step}$ grid cells in the specified direction, and the land action terminates the search.

\subsubsection{Reward}
A positive reward, $r_\textrm{dt}$, is given for each correctly discovered object, a negative reward, $r_\textrm{step}$, for each flight step, a negative reward, $r_\textrm{nfz}$, for each time the drone tries to fly outside the field and a large negative reward, $r_\textrm{crash}$, when the drone runs out of battery before landing. In addition to \citet{vanEssen2025RLPlanning}, a negative reward, $r_\textrm{nocov}$, is given when the agent flies over only previously visited grid cells, to discourage revisiting already-explored areas and reduce looping behavior.

Table~\ref{tab:sim_parameters} shows the parameter values for the MDP. Compared to \citet{vanEssen2025RLPlanning}, the initial battery level, $b_\textrm{init}$, was increased, and the negative reward for each flight step, $r_\textrm{step}$, was reduced to account for the larger field size. Additionally, the distance per flight step, $d_\textrm{step}$, was increased to compensate for the larger field size, while still maintaining sufficient overlap between consecutive waypoints.

\subsection{Network architecture and training}
\label{sec:network_architecture}
\begin{figure}[t]
   \centering
   \includegraphics[width=0.5\columnwidth]{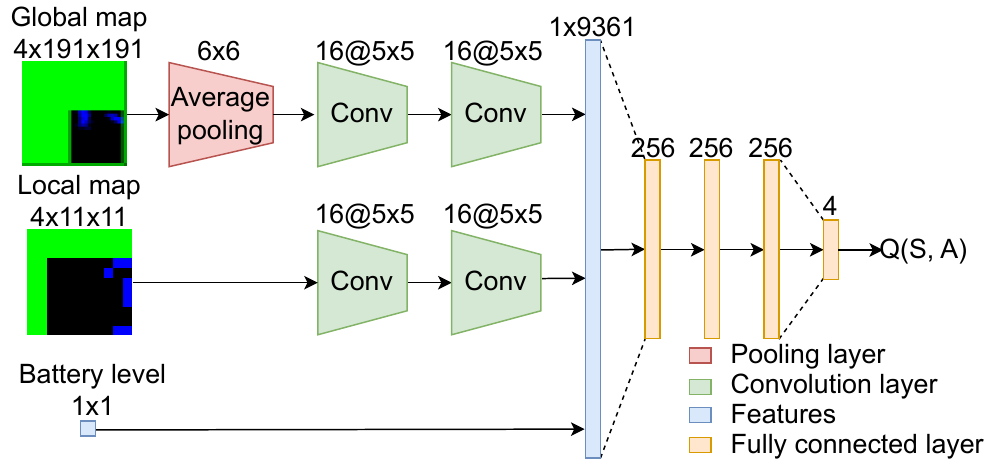}
   \caption{DQN architecture, showing the pooling, convolutional, and fully connected layers with the input size, the number of kernels and their size, and the size of the flatten layer. Modified from \citep{vanEssen2025RLPlanning}.}
   \label{fig:network_architecture}
\end{figure}

\noindent
Figure~\ref{fig:network_architecture} shows the network architecture for the RL agent used to map states to action values. To limit the number of trainable parameters, the global map is down-sampled using average pooling with a kernel size of 6 x 6. The network consists of two parallel convolutional layers that extract independent features from the local and global map. The output from both convolutional layers is flattened and concatenated with the drone's battery level. The resulting feature vector is converted to action values using three fully connected layers.

The agent was trained using deep Q-Learning (DQN) \citep{Mnih2015}, with a replay buffer size of $n_\textrm{buffer}$, a batch size of $n_\textrm{batch}$, a discount factor of $\gamma$ and a learning rate $\alpha$. Training started when 50\% of the buffer was filled, and the network was trained for $n_\textrm{steps}$ steps. The best weights were selected based on the highest average reward obtained over $n_\textrm{eval}$ evaluation simulations. Table~\ref{tab:dqn_parameters} shows the parameters used to train the RL agent. For more details about the training procedure, we refer to \citet{vanEssen2025RLPlanning}.

\begin{table}[t]
    \centering
    \small
    \caption{Training parameters for the deep Q-Learning method.}
    \label{tab:dqn_parameters}
    \begin{tabular}{lll}
        \toprule
         Symbol              & Value             & Explanation                      \\
        \midrule
         $n_\textrm{buffer}$ & 50000             & Experience replay buffer size    \\
         $n_\textrm{batch}$  & 128               & Training batch size              \\
         $\gamma$            & 0.99              & Discount factor                  \\
         $\alpha$            & $3 \cdot 10^{-5}$ & Learning rate                    \\
         $n_\textrm{steps}$  & $2 \cdot 10^7$    & Training steps                   \\
         $n_\textrm{eval}$   & 120               & Number of evaluation simulations \\
        \bottomrule
    \end{tabular}
\end{table}

\subsection{Simulation environment}
\label{sec:simulation_env}
\noindent
The path planner is trained in an abstract field simulator based on \citet{vanEssen2025RLPlanning}. The field is defined as an $M \times M$ grid, where $M \in \mathbb{N}$. The simulation itself has no physical units—grid cells only get a real-world scale during deployment through $s_\textrm{grid}$ (Section~\ref{sec:action_to_coordinate}). The number of objects in the field is sampled as $n_{\mathrm{obj}} \sim \mathcal{N}(100,30)$ and rounded to the nearest integer. These objects are distributed across $k$ clusters, where $k \sim \mathcal{N}(5,2)$. The object positions within each cluster follow a multivariate Gaussian distribution with a random mean coordinate $\boldsymbol{\mu}_i$ and covariance $\Sigma_i \in \{\Sigma_1, \Sigma_2\}$. To better reflect real-world conditions, the field border is generated as a random polygon, which can be convex or concave. The number of edges is sampled as $n_{\mathrm{border}} \sim \mathcal{N}(5,3)$ with a minimum of four edges, and the polygon is then uniformly scaled to maximize the area within the field size, so that the drone always operates in a field that uses the full $M \times M$ grid.

\begin{figure}[t]
   \centering
   \subfloat[]{\includegraphics[width=0.25\textwidth]{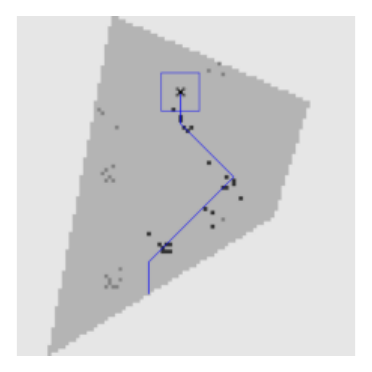}}
   \subfloat[]{\includegraphics[width=0.25\textwidth]{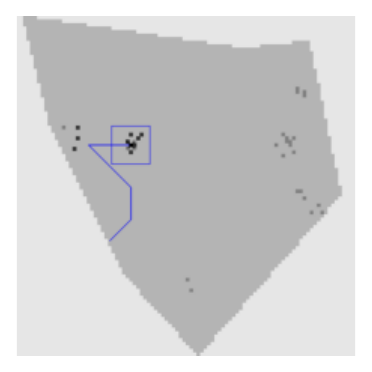}}
   \caption{Two examples of the simulation environment with the camera's field of view indicated by the blue rectangle around the drone, the flight path in blue, the detected objects in dark-gray and the not-yet detected objects in light-gray.}
  \label{fig:simulation}
\end{figure}

A drone with a top-down facing camera with a field of view of $N \times N$ grid cells, where $N\in \mathbb{N}$, is flying over the field. The drone starts at a random position at the border of the field. The observation map contains all objects detected by the simulated detection network within the camera's field of view. To simulate typical detection errors, each empty map cell has probability $p_\textrm{dt,fp}$ of being incorrectly set to 1 as a false positive, and each detection has probability $p_\textrm{dt,fn}$ of being removed from the map as a false negative. Positional errors are simulated by shifting some objects in the observation map a few pixels. Detection errors are simulated independently for each observation. The prior knowledge map is generated from the ground truth object locations, covering the entire field. Each empty grid cell has probability $p_\textrm{pk,fp}$ of being incorrectly set to 1 as a false positive, and each object location has probability $p_\textrm{pk,fn}$ of being removed from the map as a false negative. Positional errors are simulated similarly to the observation map. Errors in the prior knowledge map are sampled only once per simulation, as the prior knowledge is fixed before the flight begins. Figure~\ref{fig:simulation} shows two examples of the simulated field.

By constantly training on changing field shapes, different object distributions, and detection errors, domain randomization is achieved. This exposes the agent to a wide range of conditions during training, helping it generalize to real-world data.

Table~\ref{tab:sim_parameters} shows the parameters of the simulation environment. Compared to \citet{vanEssen2025RLPlanning}, the field size is quadrupled, the field is randomly shaped rather than a fixed square, and a larger number of object clusters are present in the field.

\begin{table}[t]
    \centering
    \small
    \caption{Parameters for the simulation environment.}
    \label{tab:sim_parameters}
    \begin{tabular}{lll}
        \toprule
         Symbol & Value & Explanation \\
        \midrule
         $M$ & 96 & Field size \\
         $n_{\mathrm{obj}}$ & $\sim \mathcal{N}(100,30)$ & Number of objects in the field \\
         $k$ & $\sim \mathcal{N}(5,2)$ & Number of clusters in the field \\
         $n_{\mathrm{border}}$ & $ \sim \mathcal{N}(5,3)$ & Number of edges of the field border (min 4) \\
         $\Sigma_1$ & [$\begin{smallmatrix} 5 & 8 \\ 8 & 15 \end{smallmatrix}]$ & Covariance matrix for distribution 1 \\
         $\Sigma_2$ & [$\begin{smallmatrix} 15 & 0 \\ 0 & 5 \end{smallmatrix}$] & Covariance matrix for distribution 2 \\
         $N$ & 11 & Field of view size of the camera \\
         $p_\textrm{dt,fp}$ & 0.05 & Probability false positive detection \\
         $p_\textrm{dt,fn}$ & 0.0001 & Probability false negative detection \\
         $p_\textrm{pk,fp}$ & 0.0005 & Probability false positive prior knowledge \\
         $p_\textrm{pk,fn}$ & 0.25 & Probability false negative prior knowledge \\
         $b_\textrm{init}$ & 200 & Initial battery level \\
         $b_\textrm{step}$ & 0.2 & Battery usage per step \\
         $d_\textrm{step}$ & 3 & Number of grid cells per step \\
         $r_\textrm{dt}$ & 1.0 & Detection reward \\
         $r_\textrm{step}$ & -0.125 & Step penalty \\
         $r_\textrm{nfz}$ & -1.0 & Penalty for entering a no-fly-zone \\
         $r_\textrm{crash}$ & -150.0 & Penalty for crashing (due to empty battery) \\
         $r_\textrm{nocov}$ & -0.5 & Penalty for not exploring new grid cells \\
        \bottomrule
    \end{tabular}
\end{table}

\subsection{Application in real world}
\label{sec:application}
\noindent
To apply the simulation-trained path planner in the real world, real-world observations must be mapped to the state-space, and the action-space must be mapped to real-world coordinates. This requires (1) training a detection network to detect objects, (2) converting real-world Universal Transverse Mercator (UTM) coordinates to the simulation grid, (3) converting detected objects into the observation map, and (4) generating real-world prior knowledge. This applies equally to the offline evaluation on orthomosaics (Section~\ref{sec:datasets}) and the real-time application on the drone (Section \ref{sec:live_application}); the two settings differ only in how the observations are obtained, either as crops from an orthomosaic or as live images from the drone's camera.

\subsubsection{Object-detection network}
\label{sec:detection_network}
\noindent
YOLOv11-m \citep{Jocher2024} was used to detect objects due to its balance between detection speed and accuracy, making it suitable for real-time inference during drone flights. The YOLOv11 detection network was trained for 250 epochs on a dataset of real-world drone images from the same grass field used to create the first four evaluation datasets (Section~\ref{sec:datasets}), taken from 12, 24, and 32m altitude. The training dataset was identical to that used in \citep{vanEssen2025AdaptivePathPlanning}, with all plant classes in that dataset merged into a single `plant' class. In total, 1618 images with 2000 plant annotations were used for training. The precision and recall on the validation set were 0.96 and 0.95 respectively.

At every flight step, the drone captures an image from a top-down perspective. This image is resized to dimensions $[I_h, I_w]^T$, and the trained YOLOv11-m network is used to detect objects. Detections with a detection confidence lower than 0.5 are discarded. For this work, $I_w=2048$ and $I_h=2048$.

\subsubsection{Coordinate conversion}
\label{sec:action_to_coordinate}
\noindent
To apply the learned adaptive path planner in the real world, the field is divided into a grid of $M \times M$ grid cells, rotated by an angle $\psi$ relative to North so that the grid can be aligned with the field's orientation. The grid coordinates, $[x^\textrm{grid},y^\textrm{grid}]^T$, of any real-world UTM coordinate, $[x^\textrm{utm},y^\textrm{utm}]^T$ (representing Easting and Northing), such as the drone's location, a detected object, or the field border, are calculated by:

\begin{equation}
    \begin{bmatrix}
        x^\textrm{local} \\
        y^\textrm{local}
    \end{bmatrix}
    =
    \begin{bmatrix}
        \cos(\psi) & -\sin(\psi) \\
        \sin(\psi) & \cos(\psi)
    \end{bmatrix}
    \cdot
    \begin{bmatrix}
         \displaystyle -\frac{y^\textrm{utm} - y^\textrm{utm}_\textrm{center}}{s_\textrm{grid}} \\
         \displaystyle\frac{x^\textrm{utm} - x^\textrm{utm}_\textrm{center}}{s_\textrm{grid}}        
    \end{bmatrix}
\end{equation}
\begin{equation}
     \begin{bmatrix}
        x^\textrm{grid} \\
        y^\textrm{grid}
    \end{bmatrix}
    =  
    \begin{bmatrix}
        x^\textrm{local} \\
        y^\textrm{local}
    \end{bmatrix} 
    + 
    \frac{M - 1}{2}
\end{equation}
\noindent
where $\psi$ is the counter-clockwise rotation of the grid relative to North, $[x^\textrm{utm}_\textrm{center}, y^\textrm{utm}_\textrm{center}]^T$ the grid center, and $s_\textrm{grid}$ the size of a single grid-cell in meters. The size of a single grid cell depends on the required resolution and area to be covered by the grid. The grid coordinates are rounded to the nearest integer.

\subsubsection{Object detections to observation map}
\label{sec:detection_to_obseration}
\noindent
The observation map had a size of $N \times N$ grid cells, corresponding to the field of view of the camera, and is independent of the drone's location. The observation map coordinates, $[x^\textrm{obs}_i,y^\textrm{obs}_i]^T$, of object $i$ are calculated by:
\begin{equation}
    x^\textrm{obs}_i = \left\lfloor \frac{N}{I_h} \cdot y^\textrm{img}_i \right\rceil, \quad
    y^\textrm{obs}_i = \left\lfloor \frac{N}{I_w} \cdot x^\textrm{img}_i \right\rceil
\end{equation}
\noindent
where $[x^\textrm{img}_i,y^\textrm{img}_i]^T$ are the image pixel coordinates of the center of object $i$, $N$ is the size of the field of view in grid cells, and $\lfloor\cdot\rceil$ denotes rounding to the nearest integer. The observation map is reset to zero at each flight step, and each grid cell that contains at least one detection is set to 1, resulting in a binary observation map consistent with the simulation.

\subsubsection{Generation of prior knowledge}
\label{sec:prior_knowledge_generation}
Prior knowledge about object locations can be derived from various sources, such as detections from previous flights or data from prior years. For this work, no historical data was available, so we generate a prior knowledge map of size $M \times M$ using a single high-altitude full-coverage flight without overlap. The field of view for the high-altitude coverage flight is set to $N_\textrm{pk} \times N_\textrm{pk} \in \mathbb{N}^2$ grid cells corresponding with a real-world size of $(N_\textrm{pk} \cdot s_\textrm{grid}) \times (N_\textrm{pk} \cdot s_\textrm{grid})$. At every waypoint, an image is taken from a top-down perspective, objects are detected using the detection network, and each prior knowledge map cell that contains at least one detection with a confidence above 0.05 is set to 1, resulting in a binary prior knowledge map. Since consecutive images do not overlap, each grid cell is covered by exactly one image. This low threshold is used on purpose: it prioritizes recall over precision by also including uncertain detections. Since the RL agent is trained with noisy prior knowledge, completely missing an object is more harmful than including a false positive. For this work, $N_\textrm{pk}=24$, corresponding to a total of $(M/N_\textrm{pk})^2 = (96/24)^2 = 16$ images for the whole field.

\subsection{Orthomosaic datasets}
\label{sec:datasets}
\begin{figure}[t]
   \centering
    \includegraphics[width=0.3\columnwidth]{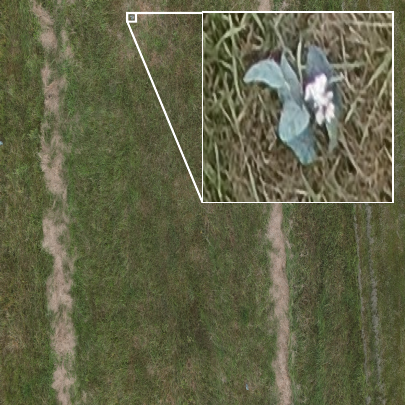}
   \caption{Example of the artificial plants in the grass field.}
   \label{fig:example_annotation}
\end{figure}

\noindent
To allow repeatable experiments on the same data, the RL agent is evaluated on seven offline, real-world image datasets. Each dataset consists of a high-resolution orthomosaic — a geometrically corrected aerial image composed of many overlapping drone images, providing a detailed and geo-referenced top-down view of the field — in the UTM coordinate system. The orthomosaic contains a grass field with commercially available plastic plants to simulate a weed detection application (see Figure~\ref{fig:example_annotation}). These artificial plants were used because their positions can be precisely measured and they can be reused across experiments. The plants are distributed in clusters, and their ground-truth locations were measured using a Topcon HiPer SR RTK-GNSS receiver for the first four datasets and an Emlid Reach RS3 RTK-GNSS receiver for the last three datasets. Images from the field were taken using a DJI M300 drone with a Zenmuse P1 RGB camera at 12m altitude with 70\% side overlap and 80\% front overlap for the first four datasets, and using a DJI M350 drone with identical overlap settings for the last three datasets. The orthomosaics were made using Agisoft Metashape \citep{Agisoft2023} with the high-quality settings. Note that the field shape and area slightly differ between the datasets due to the availability of some parts of the field.

\subsection{Live real-world application}
\label{sec:live_application}
\begin{figure}[t]
   \centering
   \includegraphics[width=0.4\textwidth]{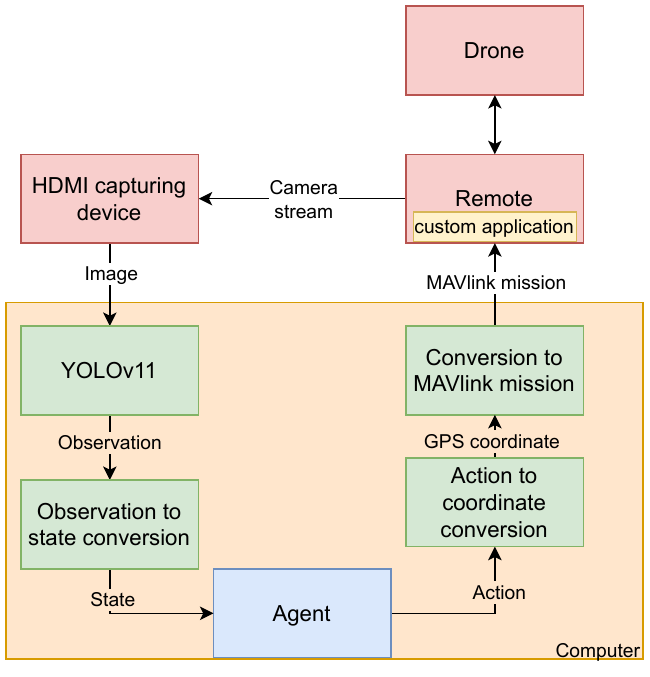}
   \caption{Deployment of the RL agent on the real drone using an HDMI capturing device, YOLOv11 and a custom application on the drone's remote controller that executes MAVLink missions.}
  \label{fig:drone_mavlink}
\end{figure}

\noindent
To assess the feasibility of the simulation-trained RL policy in real-world conditions, we deployed it on a DJI M350 drone with a Zenmuse P1 camera. Figure~\ref{fig:drone_mavlink} illustrates the deployment of the RL agent on the drone. The agent's actions are converted to GPS coordinates (Section~\ref{sec:action_to_coordinate}) and then translated into the MAVLink communication protocol. These messages are sent to the drone's remote controller via Ethernet. On the drone's remote controller, we developed a custom flight application using the DJI MSDK, which converts the MAVLink missions into DJI missions and executes them on the drone's onboard flight controller. The M350 receives real-time kinematic (RTK) correction data over NTRIP, enabling centimeter-level positioning accuracy during flight. The camera images are captured from the remote using an HDMI capturing device to ensure a stable connection. The weeds are detected in the image using YOLOv11 (Section~\ref{sec:detection_network}) and converted to the state representation (Section~\ref{sec:detection_to_obseration}). The custom flight application is available at \url{https://github.com/wur-abe/mavdrone}.

\subsection{Levels of realism}
\label{sec:levels_of_realism}
To evaluate the gap between simulation and reality for the simulation-trained RL policy, the path planner was evaluated on six levels of realism. Each level replaces one or more additional components: object positions, prior knowledge, observations, or the flight controller, with its real-world counterpart, as shown in Table~\ref{tab:realism_levels}. \textit{Level 1} corresponds to the original training simulation environment as defined in Section~\ref{sec:simulation_env}. \textit{Level 2} uses the same environment but replaces the randomly generated object positions with the real-world weed positions from the seven datasets, with the corresponding orthomosaic rotated by an angle $\psi$ to align it with the simulation grid (Section~\ref{sec:action_to_coordinate}). \textit{Level 3} additionally replaces the simulated prior knowledge with a prior knowledge map generated from a high-resolution orthomosaic of the field (Section~\ref{sec:datasets}). \textit{Level 4} additionally replaces the simulated observation with observations derived from the same orthomosaic: at each flight step, the camera image is simulated by cropping the orthomosaic at the drone's location, and the trained YOLOv11 detection network (Section~\ref{sec:detection_network}) is applied to this crop to produce the observation map. The flight controller remains simulated at this level. \textit{Level 5} additionally replaces the orthomosaic-generated prior knowledge with real prior knowledge collected during the actual drone flight (Section~\ref{sec:prior_knowledge_generation}), while observations are still derived from the orthomosaic. \textit{Level 6} corresponds to the full real-world application (Section~\ref{sec:live_application}), where the drone flies the field using the real flight controller, capturing live camera images at each step.

\begin{table}[]
    \centering
    \small
    \caption{Definition of the six realism levels used to evaluate the simulation-to-reality gap. Each level replaces one additional component with its real-world counterpart. `S' denotes simulated, `G' denotes generated from a high-resolution orthomosaic, and `R' denotes real-world data collected and processed during the drone flight.}
    \label{tab:realism_levels}
    \begin{tabular}{ccccc}
        \toprule
         Level & Object positions & Prior knowledge & Observation & Flight controller \\
        \midrule
         1     & S              & S               & S           & S                 \\ 
         2     & R              & S               & S           & S                 \\ 
         3     & R              & G               & S           & S                 \\ 
         4     & R              & G               & G           & S                 \\ 
         5     & R              & R               & G           & S                 \\ 
         6     & R              & R               & R           & R                 \\ 
        \bottomrule
    \end{tabular}
\end{table}

\subsection{Performance metrics}
\label{sec:performance_metrics}
\noindent
The RL agent was evaluated using two types of metrics: flight path length and detection performance. The flight path length is expressed in number of flight steps and is used to assess the efficiency of the path planner. Detection performance was evaluated at three levels: field, image, and prior knowledge, which respectively reflect the overall flight performance, the quality of the local map, and the quality of the global map. At the \textit{field level}, precision and recall were computed based on the total number of detected objects compared to the ground truth. An object was considered a true positive if its detected position was within a distance of $0.99 \times s_\textrm{grid}$ from the ground-truth location. The factor 0.99 accounts for small measurement errors; without it, even slight deviations of a few centimeters could place a detection in a neighboring grid cell, resulting in a false positive and a false negative despite the detection being essentially correct. Detections without a corresponding ground-truth object within this threshold were counted as false positives, while ground-truth objects without nearby detections were counted as false negatives. At the \textit{image level}, precision and recall were calculated for each image individually, using the same distance threshold as at the field level. True positives, false positives, and false negatives were aggregated across images to compute overall performance metrics. At the \textit{prior knowledge level}, precision and recall were evaluated based on the prior knowledge map by comparing the prior knowledge map to the ground truth object locations. As a baseline, the RL path planner was compared to a traditional full coverage planner, with the flight path generated using Fields2Cover \citep{Mier2023}, at the field and image levels, using the camera's field of view, $N$, as the row width and waypoint distance to create no overlap.

\subsection{Experiments}
\label{sec:experiments}
\noindent
To evaluate the efficiency and generalization of the RL agent to real-world data, two experiments were performed. In the first experiment, the planner is evaluated offline without a real drone on realism levels 1--4 (Section~\ref{sec:levels_of_realism}), allowing us to quantify the simulation-to-reality gap. The second experiment demonstrates the feasibility of deploying the simulation-trained RL policy on a real drone (level~6), and additionally evaluates level~5 using prior knowledge collected during the actual drone flight for better comparison with the offline evaluation.

\subsubsection{Experiment 1: Simulation-to-Reality gap}
Since levels 1--4 do not require a real drone, all evaluations in this experiment were conducted offline. Level~1 was evaluated using 1000 randomly generated fields. Levels 2--4 were evaluated using the seven orthomosaic datasets described in Section~\ref{sec:datasets}. To increase repetitions, each orthomosaic was evaluated four times with a different grid rotation $\psi$, generating a unique state-space for the RL agent, resulting in 28 repetitions in total. We compared precision, recall, and flight path length for both the RL agent and the baseline full coverage planner at the field, image, and prior knowledge levels.

\subsubsection{Experiment 2: Real-world drone flight}
Because real-world drone flights are time-consuming and logistically challenging, experiment~2 does not include repeated trials like experiment~1. Instead, it serves as a proof-of-concept demonstration of how the RL policy performs in real conditions and to highlight differences between real-world deployment and simulation. Two drone flights were made on the same field with a different grid rotation, $\psi$. Prior knowledge was collected as described in Section~\ref{sec:prior_knowledge_generation}. The ground truth weed positions were measured using an Emlid Reach RS3 RTK-GNSS receiver. An orthomosaic was made of the same field and weed distribution used in the drone flights, allowing the real-world results (level~6) to be directly compared to the orthomosaic-based evaluation at levels 2--5.

\section{Results}
\subsection{Experiment 1: Simulation-to-Reality gap}
\begin{figure}[t]
  \centering
  \subfloat[]{\includegraphics[width=0.45\linewidth]{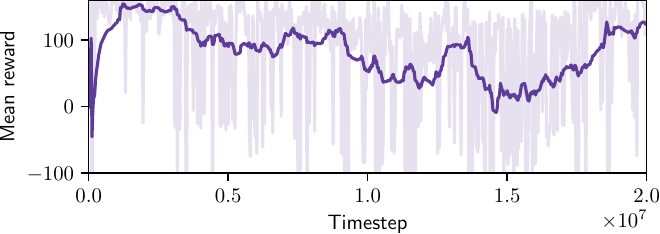}}
  \subfloat[]{\includegraphics[width=0.45\linewidth]{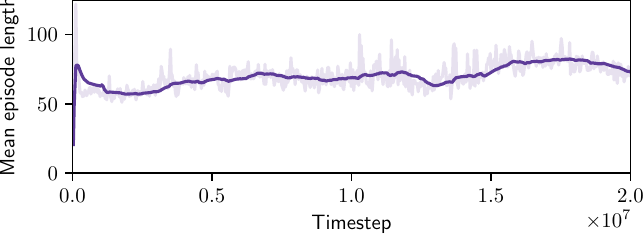}}
  \caption{Mean reward (a) and mean episode length (b) of the evaluation simulations during training of the DQN agent. The line shows the moving average over a 1M-timestep window (partial window for timesteps < 1M).}
  \label{fig:training_metrics}
\end{figure}

\noindent
Figure~\ref{fig:training_metrics} shows the mean reward and mean episode length of the evaluation simulations during training of the DQN agent at different training timesteps. The DQN agent yielded the highest reward after around 1.5 million training steps. However, as training continued, the agent also learned to increase the episode length while also increasing the mean reward, indicating that the agent learned to find weeds that are farther away from the starting position. 

\begin{figure}[t]
   \centering
   \subfloat[]{\includegraphics[width=0.47\columnwidth]{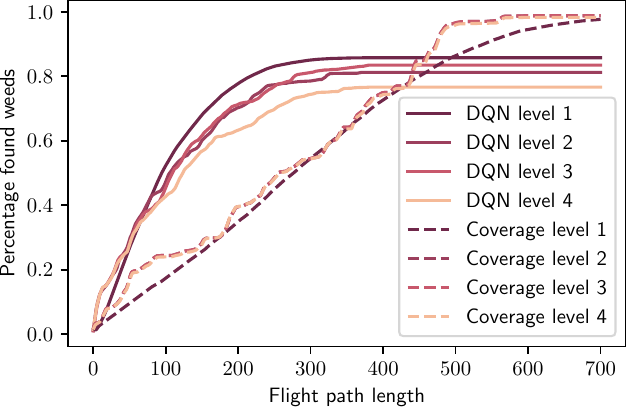}}
   \hfill
   \subfloat[]{\includegraphics[width=0.47\columnwidth]{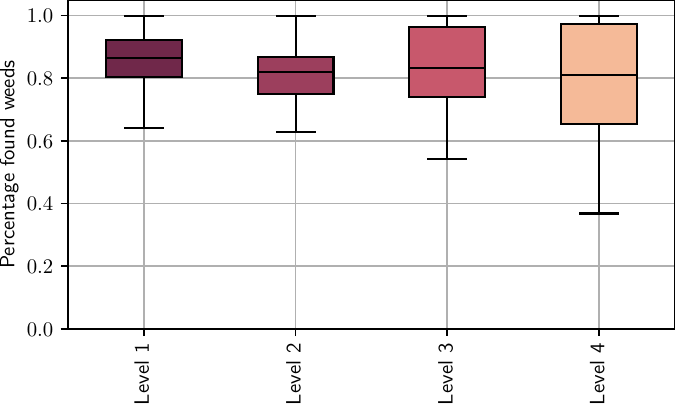}}
   \caption{(a) Effect of flight path length on the recall at field level for the RL agent (DQN) and the baseline full coverage planner across the different realism levels. Note that the lines for 'Coverage level 2' and 'Coverage level 3' are equal. (b) Boxplot of the recall at the end of the flight for the RL agent for the different realism levels, showing the distribution and variance in percentage of found weeds.}
   \label{fig:realism_levels_result}
\end{figure}

Figure~\ref{fig:realism_levels_result}a shows the effect of flight path length on recall for both the RL agent and the baseline full coverage planner across the four realism levels. Overall, the recall of the RL agent decreases as realism increases. However, up to approximately 450 flight steps, the RL agent discovered weeds faster than the full coverage planner. The performance difference between levels 1 and 2, which results from introducing weed distributions derived from real-world orthomosaic data, is approximately 2\%, suggesting that the RL agent generalizes well across different weed distributions. The transition from level 2 to level 3, where prior knowledge generated from orthomosaic data was introduced, leads to a small improvement in performance. Note that the full coverage planner does not utilize prior knowledge, and therefore its performance remains unchanged between levels 2 and 3. The relatively small performance decrease of around 5\% from level 3 to level 4 indicates that the detection network was accurately simulated, and that the agent generalizes well to the outputs of a real detection network. The full coverage planner did not find all weeds on the real-world dataset (level 4) due to detection errors. Figure~\ref{fig:realism_levels_result}b shows the recall at the end of the flight per realism level. Levels 1 and 2 showed a similar variance, with a slightly higher median at level 1. At levels 3 and 4, the variance increased, with more cases achieving a low recall, though the majority of cases still achieved a recall above 0.7.

Table~\ref{tab:recall_results_per_step} shows the recall for the RL agent and the baseline full coverage planner at 150, 300, 450, and 600 flight steps across the different realism levels. Up until 300 flight steps, the RL agent had a significantly higher recall than the full coverage planner, showing that the RL agent found more weeds at that timestep. Already at 150 flight steps, the RL agent found more than 57\%-68\% of the weeds on average, compared to only 26\%-28\% for the full coverage planner. After 450 flight steps, the full coverage planner had found more weeds than the RL agent, except when using fully simulated fields (level 1).

\begin{table}[t]
    \centering
    \small
    \caption{Recall values at different flight steps for the RL agent (DQN) and the baseline full coverage planner. The values show the mean $\pm$ the standard deviation. Values marked with a '*' have a significant ($\alpha=0.001$, Welch’s t-test) higher recall than the full coverage planner.}
    \label{tab:recall_results_per_step}
    \begin{tabular}{@{}c@{}cllll}
        \toprule
         \multirow{2}{*}{} & \multirow{2}{*}{\makecell{Realism\\level}} & \multirow{2}{*}{150 steps} & \multirow{2}{*}{300 steps} & \multirow{2}{*}{450 steps} & \multirow{2}{*}{600 steps} \\
         & & & & & \\
        \midrule
         \multirow[origin=c]{4}{*}{\rotatebox{90}{DQN}}  
         & 1 & 0.68 ± 0.15* & 0.85 ± 0.09* & 0.86 ± 0.08* & 0.86 ± 0.08 \\
         & 2 & 0.66 ± 0.18* & 0.80 ± 0.10* & 0.81 ± 0.09  & 0.81 ± 0.09 \\
         & 3 & 0.59 ± 0.24* & 0.80 ± 0.17* & 0.83 ± 0.15  & 0.83 ± 0.15 \\
         & 4 & 0.57 ± 0.26* & 0.74 ± 0.21* & 0.77 ± 0.22  & 0.77 ± 0.22 \\
        \midrule
         \multirow[origin=c]{4}{*}{\rotatebox{90}{Coverage}}  
         & 1 & 0.26 ± 0.19  & 0.55 ± 0.23  & 0.80 ± 0.19  & 0.95 ± 0.10 \\
         & 2 & 0.28 ± 0.14  & 0.54 ± 0.19  & 0.85 ± 0.09  & 0.99 ± 0.03 \\
         & 3 & 0.28 ± 0.14  & 0.54 ± 0.19  & 0.85 ± 0.09  & 0.99 ± 0.03 \\
         & 4 & 0.28 ± 0.15  & 0.54 ± 0.19  & 0.84 ± 0.09  & 0.98 ± 0.04 \\
        \bottomrule
    \end{tabular}
\end{table}

Table~\ref{tab:pr_field_image_pk_level} presents the flight path length as well as the precision and recall at field, image, and prior knowledge levels for both the RL agent and the full coverage planner across different realism levels. For all realism levels, the RL agent achieves significantly shorter flight paths than the full coverage planner, with reductions of approximately 57, 58, 39, and 38 percentage points at levels 1 to 4, respectively. At both field and image levels, the DQN agent consistently has higher precision than the full coverage planner, which can be caused by the shorter flight paths. Consequently, the RL agent captures fewer images and thereby has a lower probability of producing false positive detections. With the longer flight path, the full coverage planner was able to detect almost all weeds in the field (with a field level recall of 0.98–0.99 across realism levels), whereas the RL agent showed a lower recall (approximately 0.77–0.85). At level~1 (pure simulation), this amounted to a 57\% shorter flight path at the cost of a 13\% lower field-level recall (0.85 versus 0.98); at level~4 (real-world orthomosaic data), the RL agent achieved a 38\% shorter flight path at the cost of a 21\% lower recall (0.77 versus 0.98).

With increasing realism, the performance of the RL agent shows a gradual decline in both precision and recall, showing the added complexity at each realism level. The slight improvement in field-level recall between realism levels 2 and 3 can be attributed to a marginally higher prior knowledge recall derived from orthomosaic data compared to simulated data. In contrast, the decrease in both field-level precision and recall between realism levels 3 and 4 is likely caused by reduced image-level performance, resulting from the use of a real detection network on the orthomosaic data instead of the simulated detection network output.

\begin{table}[t]
    \centering
    \small
    \addtolength{\leftskip} {-2cm}
    \addtolength{\rightskip}{-2cm}
    \caption{Flight path length and the precision and recall values at field, image and prior knowledge level for the RL agent (DQN) and the full coverage planner across different realism levels. The values show the mean ± the standard deviation. Flight paths marked with a '*' are significantly ($\alpha=0.001$, Welch’s t-test) shorter than the corresponding coverage flight path at the same realism level.}
    \label{tab:pr_field_image_pk_level}
    \begin{tabular}{@{}c@{}cccccccc}
        \toprule
         \multirow{2}{*}{} & \multirow{2}{*}{\makecell{Realism\\level}} & \multirow{2}{*}{\makecell{Flight path\\length}} & \multicolumn{2}{c}{Field level} & \multicolumn{2}{c}{Image level} & \multicolumn{2}{c}{Prior knowledge level}  \\
       \cmidrule(lr{2pt}){4-5} % 2pt shorter
       \cmidrule(lr{2pt}){6-7}
       \cmidrule(lr{2pt}){8-9}
        &   &         & Precision & Recall    & Precision & Recall    & Precision & Recall    \\
       \midrule
        \multirow[origin=c]{4}{*}{\rotatebox{90}{DQN}}  
        & 1 & 239±57* & 0.92±0.04 & 0.85±0.08 & 0.97±0.02 & 0.97±0.02 & 0.48±0.11 & 0.26±0.06 \\
        & 2 & 251±64* & 0.92±0.03 & 0.81±0.08 & 0.96±0.04 & 0.98±0.01 & 0.44±0.03 & 0.24±0.02 \\
        & 3 & 369±17* & 0.86±0.04 & 0.83±0.13 & 0.95±0.03 & 0.97±0.01 & 0.27±0.06 & 0.26±0.06 \\
        & 4 & 375±11* & 0.96±0.05 & 0.77±0.22 & 0.93±0.09 & 0.96±0.05 & 0.26±0.06 & 0.27±0.06 \\
       \midrule
        \multirow[origin=c]{4}{*}{\rotatebox{90}{Coverage}}  
        & 1 & 558±103 & 0.85±0.06 & 0.98±0.05 & 0.94±0.06 & 0.97±0.01 &           &           \\
        & 2 & 605±92  & 0.90±0.04 & 0.99±0.04 & 0.92±0.07 & 0.98±0.01 &           &           \\
        & 3 & 605±92  & 0.89±0.03 & 0.99±0.04 & 0.92±0.07 & 0.98±0.01 &           &           \\
        & 4 & 605±92  & 0.94±0.09 & 0.98±0.04 & 0.87±0.12 & 0.96±0.03 &           &           \\
       \bottomrule
    \end{tabular}
\end{table}

\begin{figure}[t]
   \centering
   \subfloat[]{\includegraphics[width=0.24\textwidth]{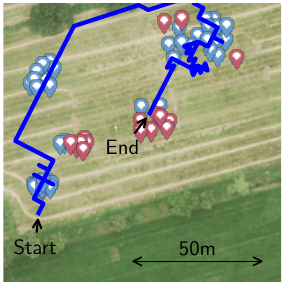}}
   \subfloat[]{\includegraphics[width=0.24\textwidth]{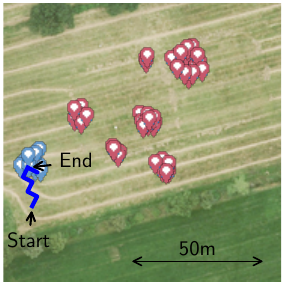}}
   \subfloat[]{\includegraphics[width=0.24\textwidth]{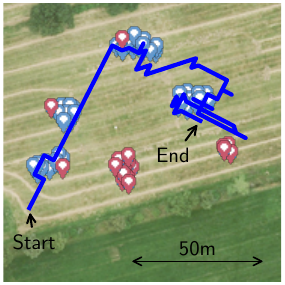}}
   \subfloat[]{\includegraphics[width=0.24\textwidth]{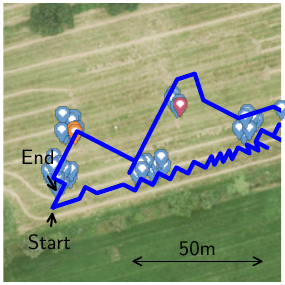}}
   \\
   \subfloat[]{\includegraphics[width=0.24\textwidth]{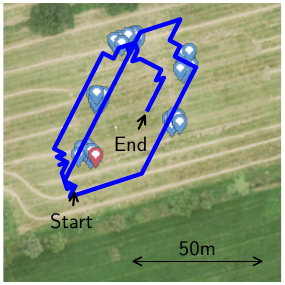}}
   \subfloat[]{\includegraphics[width=0.24\textwidth]{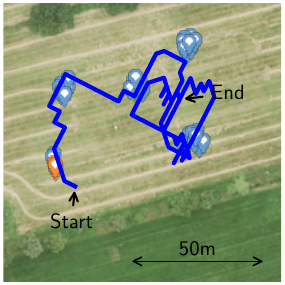}}
   \subfloat[]{\includegraphics[width=0.24\textwidth]{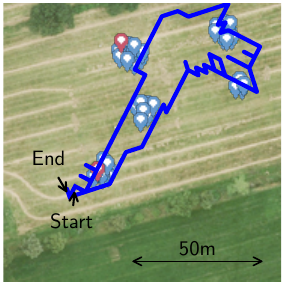}}
   \caption{Some flight paths for the RL agent on the seven real-world datasets. Blue, orange, and red markers indicate true positive, false positive, and false negative markers, respectively. The start and end points of each flight path are indicated with arrows.}
   \label{fig:flight_paths_orthomosaic_results}
\end{figure}

Figure~\ref{fig:flight_paths_orthomosaic_results} shows some flights for the RL agent on the seven real-world datasets. In most cases, the agent successfully located the majority of weed clusters, though not all weeds within each cluster were detected. Figure~\ref{fig:flight_paths_orthomosaic_results}b shows a less successful case, where the agent terminated after discovering only one cluster, missing the others. Although the agent generally discovers weeds quickly, the flight paths were not always efficient: the agent sometimes revisited already-covered areas and in several cases continued flying after most weeds had already been found, delaying the land action and thereby increasing the flight path length unnecessarily.

\subsection{Experiment 2: Real-world drone flight}
\begin{figure}[t]
   \centering
   \setlength{\tabcolsep}{1pt}
   \begin{tabular}{cccccc}
      & \small Level 2 & \small Level 3 & \small Level 4 & \small Level 5 & \small Level 6 \\
      \raisebox{1.2cm}{\rotatebox{90}{\small Trial 1}} &
      \subfloat[]{\includegraphics[width=0.18\textwidth]{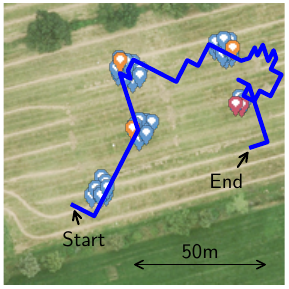}} &
      \subfloat[]{\includegraphics[width=0.18\textwidth]{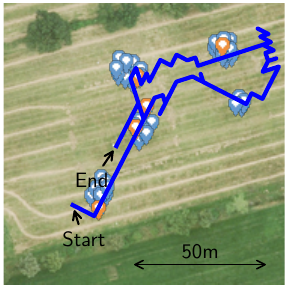}} &
      \subfloat[]{\includegraphics[width=0.18\textwidth]{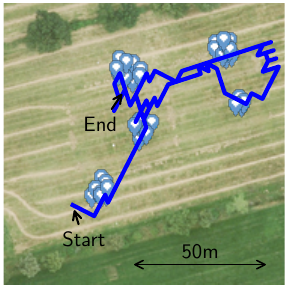}} &
      \subfloat[]{\includegraphics[width=0.18\textwidth]{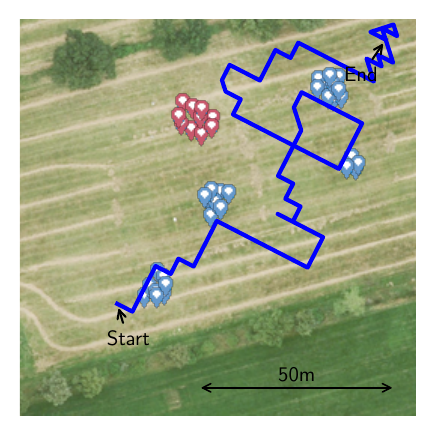}} &
      \subfloat[]{\includegraphics[width=0.18\textwidth]{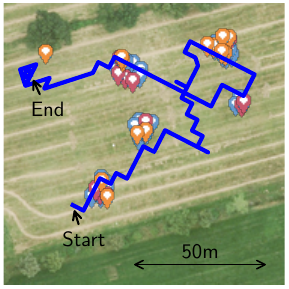}} \\
      \raisebox{1.2cm}{\rotatebox{90}{\small Trial 2}} &
      \subfloat[]{\includegraphics[width=0.18\textwidth]{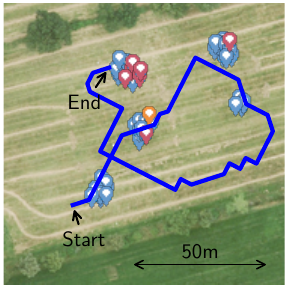}} &
      \subfloat[]{\includegraphics[width=0.18\textwidth]{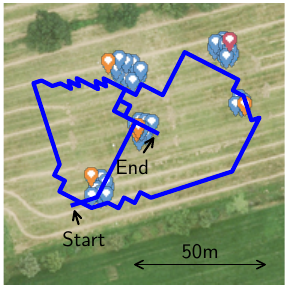}} &
      \subfloat[]{\includegraphics[width=0.18\textwidth]{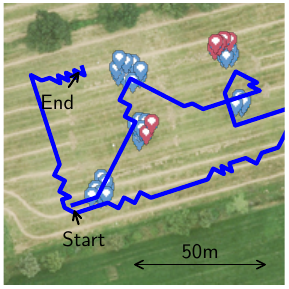}} &
      \subfloat[]{\includegraphics[width=0.18\textwidth]{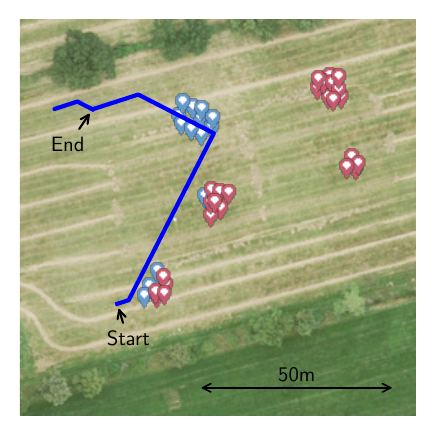}} &
      \subfloat[]{\includegraphics[width=0.18\textwidth]{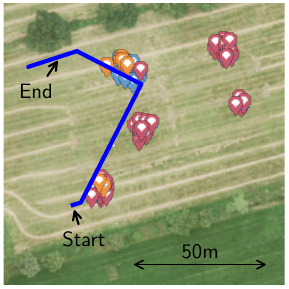}} \\
   \end{tabular}
   \caption{Flight paths for trial 1 (a--e) and trial 2 (f--j) across realism levels 2 (a, f), 3 (b, g), 4 (c, h), 5 (d, i), and 6 (e, j). Level 6 corresponds to the real-world drone flight. Blue, orange, and red markers indicate true positive, false positive, and false negative detections, respectively. The start and end points of each flight path are indicated with arrows.}
   \label{fig:flight_paths_real}
\end{figure}

\noindent
Figure~\ref{fig:flight_paths_real} shows the flight paths (\textit{level 6}) for the two real-world validation trials and the corresponding flight paths for realism \textit{levels 2--5}. In real-world trial~1, the agent found 73\% of the weeds within 91 flight steps, successfully detecting all weed clusters; in trial~2, it found 23\% of the weeds within 93 flight steps, missing most clusters. Both trials were manually terminated after these steps, as the drone repeatedly executed the same sequence of actions without flying to a new position; similar looping behavior was also observed in experiment~1, as shown in Figure~\ref{fig:flight_paths_orthomosaic_results}. The flight trajectories generated under realism levels~2--4 are somewhat comparable to one another, showing a similar overall pattern, whereas the two real-world validation trials show a different pattern compared to the realism levels. The largest change between levels occurs between levels~4 and~5, the only difference being that level~5 uses real-world prior knowledge instead of the orthomosaic-generated prior knowledge used at level~4. In experiment~2, the flight paths of realism level~5 (Figure~\ref{fig:flight_paths_real}d, i) are visually most comparable to the real-world flights (Figure~\ref{fig:flight_paths_real}e, j; level~6).

Table~\ref{tab:pr_field_image_pk_level_real_world} shows real-world field test results at field, image, and prior knowledge levels compared to realism levels~2–5. Image-level precision and recall were lower in the real-world flights (level~6) than at level~5, resulting in lower field-level performance. Further analysis showed that the poor performance is explained by the failure to acquire a usable prior knowledge map. In Trial~1, prior knowledge precision and recall fell to 0.03 and 0.05 respectively, compared to 0.24 and 0.23 for the prior knowledge generated from orthomosaic data (levels~2 and 3). In Trial~2, both metrics dropped to 0.00, while levels~2 and 3 show a precision and recall of 0.33 and 0.30. Realism level~5, using real-world prior knowledge with the orthomosaic data, shows similar field-level recall as the real-world drone flight, indicating that inaccurate prior knowledge, rather than the real-world flight conditions themselves, was the main cause. Many false positives appeared along a tree row at the field edge, outside the orthomosaic but visible in the real flights. Swapping realism level~4 prior knowledge with the real-world version (level 5) produced a similar drop, confirming prior knowledge errors as the dominant source of performance loss.

\begin{table}[t]
    \centering
    \small
    \addtolength{\leftskip} {-2cm}
    \addtolength{\rightskip}{-2cm}
    \caption{Flight path length; field-level precision and recall; image-level precision and recall; and prior-knowledge-level precision and recall for the two real-world drone flights (level~6), along with the corresponding results from the orthomosaic and the simulation environments of the same field and weed distribution. Note that the real drone flights were terminated prematurely; therefore, no flight path length could be calculated.}
    \label{tab:pr_field_image_pk_level_real_world}
    \begin{tabular}{@{}c@{}cccccccc}
    \toprule
     \multirow{2}{*}{} & \multirow{2}{*}{\makecell{Realism\\level}} & \multirow{2}{*}{\makecell{Flight path\\length}} & \multicolumn{2}{c}{Field level} & \multicolumn{2}{c}{Image level} & \multicolumn{2}{c}{Prior knowledge level}  \\
    \cmidrule(lr{2pt}){4-5} % 2pt shorter
    \cmidrule(lr{2pt}){6-7}
    \cmidrule(lr{2pt}){8-9}
      &      &        & Precision & Recall & Precision & Recall & Precision & Recall \\
    \midrule
     \multirow[origin=c]{4}{*}{\rotatebox{90}{Trial 1}}  
      & 2    & 239.95 & 0.93      & 0.95   & 0.98      & 0.99   & 0.44      & 0.25   \\
      & 3    & 355.10 & 0.89      & 1.00   & 0.97      & 0.97   & 0.24      & 0.23   \\
      & 4    & 375.38 & 1.00      & 1.00   & 0.97      & 0.99   & 0.24      & 0.23   \\
      & 5    & 378.93 & 1.00      & 0.75   & 0.99      & 0.99   & 0.03      & 0.05   \\
      & 6    & -      & 0.69      & 0.73   & 0.72      & 0.56   & 0.03      & 0.05   \\
    \midrule 
     \multirow[origin=c]{4}{*}{\rotatebox{90}{Trial 2}}
      & 2    & 235.19 & 0.91      & 0.83   & 0.99      & 1.00   & 0.41      & 0.23   \\
      & 3    & 376.53 & 0.89      & 0.98   & 0.97      & 0.98   & 0.33      & 0.30   \\
      & 4    & 376.40 & 1.00      & 0.83   & 0.99      & 0.98   & 0.33      & 0.30   \\
      & 5    & 378.34 & 1.00      & 0.35   & 1.00      & 1.00   & 0.00      & 0.00   \\
      & 6    & -      & 0.53      & 0.23   & 0.70      & 0.54   & 0.00      & 0.00   \\
    \bottomrule
    \end{tabular}
\end{table}

\section{Discussion}
\noindent
The RL agent shows that learning-based adaptive path planning can significantly improve efficiency in precision agriculture over full coverage planners when not all weeds need to be detected, as is the case in weed detection. This is because the DQN agent is not designed for completeness: it deliberately trades recall for a shorter flight path, so the maximum recall it can achieve is bounded by the accuracy of the detection network. On real-world datasets, it achieved 38\% shorter flight paths compared to a full coverage planner, but this came at a 21\% lower recall.

\subsection{Simulation-to-reality gap}
Since each realism level replaces one or more simulated components with its real-world counterpart (Table~\ref{tab:realism_levels}), the resulting change in performance can be discussed component by component, using the offline results from experiment~1 (levels~1--4) and the real-world flight results from experiment~2 (levels~5--6).

\subsubsection{Object distribution}
Between levels~1 and~2 (experiment~1), the object distribution changed from the fully simulated distribution used during training to a real-world distribution, reducing field-level recall from 0.85 to 0.81, a 4 percentage-point drop (Table~\ref{tab:pr_field_image_pk_level}). Note that this real-world distribution was still based on artificial plants placed in clusters in the field rather than naturally occurring weeds, though their positions were physically measured and photographed like any other real-world object. In simulation, weed clusters were spread across the whole field, whereas the real field concentrated weeds in specific regions \citep{Cardina1997, Dessaint1991}, leaving parts of the simulated state space unused during training. This suggests too little randomness in cluster placement and within-cluster distribution during training. Sampling a wider range of cluster counts and covariances ($\Sigma_1$, $\Sigma_2$), or augmenting training samples by rotating the grid-based state space (with actions rotated accordingly, e.g., a 90° rotation turning `fly north' into `fly east'), could reduce this gap.

\subsubsection{Prior knowledge}
Between levels~2 and~3 (experiment~1), replacing the simulated prior knowledge with a map generated from the real orthomosaic gave a small recall improvement rather than a drop, indicating the simulated prior knowledge already approximated real-world prior knowledge reasonably well. In experiment~2, however, replacing this orthomosaic-generated prior knowledge with real-world prior knowledge collected during the actual flight (levels~4 to~5) caused the largest drop in the whole pipeline: field-level recall fell from 1.00/0.83 to 0.75/0.35 for trials~1 and~2, respectively, as prior knowledge recall dropped from 0.23/0.30 to 0.05/0.00 (Table~\ref{tab:pr_field_image_pk_level_real_world}). This was also observed during the real-world drone flights (level 6), where the RL agent found 73\% and 23\% of the weeds in trials~1 and~2, with prior knowledge recall of 5\% and 0\% respectively. This indicates that prior knowledge does not need to be highly accurate, but that a recall of 0\% clearly leads to performance issues. The prior knowledge was generated using a single high-altitude coverage flight at 35~m, resulting in 16 images; at this altitude, a low recall can be expected given the camera setup used. Lowering the altitude could improve prior knowledge quality but would increase flight path length, reducing the RL agent's efficiency. Alternatively, detections from previous flights could be reused as prior knowledge. This would remove the need for a dedicated coverage flight and could provide a better prior map than a single high-altitude flight.

\subsubsection{Detection network}
Between levels~3 and~4 (experiment~1), replacing the simulated detector with the real YOLOv11 network reduced field-level recall from 0.83 to 0.77, a 6 percentage-point drop. The real detection network showed a 1--2\% lower image-level recall than the simulated one, with higher variance, so some weeds were likely missed entirely. Increasing the simulated detection errors during training could improve the agent's robustness to detection errors, though consistently missed weeds would still require improving the detection network itself.

\subsubsection{Flight controller}
Between levels~5 and~6 (experiment~2), replacing the simulated flight controller with the real drone isolated the effect of physically flying, since both levels use identical real-world prior knowledge. Image-level precision and recall dropped sharply, from 0.99--1.00 at level~5 to 0.70--0.72 and 0.54--0.56 in the real flights, likely due to the lower quality of the live HDMI video feed and varying lighting conditions compared to the carefully stitched orthomosaic. Field-level recall was affected less (0.75/0.35 at level~5 versus 0.73/0.23 in the real flights), since a weed only needs to be detected once across the flight.

\subsection{Flight path efficiency}
In both experiments, the agent sometimes got stuck repeating the same two opposite actions, such as flying north and south alternately, without progressing toward new areas. \Citet{vanEssen2025RLPlanning} observed similar looping behavior and found that the learned land action largely resolved it, as the agent could terminate the search when further exploration was unprofitable. However, in the real-world trials, looping still occurred and required manual termination. Although the coverage map in the state space should inform the agent that it has already visited an area, this was apparently insufficient to prevent looping, suggesting that the coverage penalty $r_\textrm{nocov}$ was too low during training to sufficiently discourage revisiting already-covered areas. Increasing $r_\textrm{nocov}$ during training could reduce this behavior. Another practical solution could be to detect looping and trigger a forced landing when the same sequence of actions is repeated.

In addition to this looping behavior, the flight paths were also not always efficient: the agent sometimes revisited already-covered areas, continued flying well after most weeds had already been found, delaying the land action, and, more generally, did not always follow the shortest route to a chosen region, since the agent selects discrete single-step actions without any explicit notion of geometric optimality. Together, these behaviors unnecessarily increased the flight path length (Figure~\ref{fig:flight_paths_orthomosaic_results}) and indicate that the trained policy did not always correctly balance the exploration-exploitation trade-off between continuing to search and landing. As with the looping behavior, increasing $r_\textrm{nocov}$ could further discourage the revisiting of already-covered areas, while increasing the step penalty $r_\textrm{step}$ relative to the detection reward $r_\textrm{dt}$ during training could shift the balance toward earlier landing, though at the risk of terminating the search too early and missing weeds. A more structural solution to the routing inefficiency would be a hybrid approach, where the RL agent selects which regions to visit next based on prior knowledge and current detections, while a classical coverage path planner, such as Fields2Cover \citep{Mier2023}, computes the efficient route between these regions. Alternatively, a hierarchical, multi-resolution strategy could be used, where the drone first flies a coverage path at high altitude for a broad overview of the field, and the RL agent subsequently determines which sub-regions warrant closer inspection at a lower altitude, similar to the multi-resolution adaptive planning approach of \citet{Stache2023}.

\subsection{Comparison with rule-based planning}
\Citet{vanEssen2025AdaptivePathPlanning} proposed a rule-based adaptive planner for the same weed-detection task, evaluated on the first four datasets, achieving a 37\% shorter flight path than a full coverage planner at only a 2\% lower F1-score, i.e., higher precision and recall than the RL agent at a comparable flight path length. Note that the rule-based planner was not evaluated on the last three datasets. \citet{Stache2023} reported similar efficiency gains from a non-learned adaptive planner outside weed detection, proposing a decision-function-based approach that reduced flight altitude, and therefore flight time, only in areas identified as semantically interesting, improving data-collection efficiency for crop monitoring without relying on a learned policy. However, rule-based and other non-learned adaptive planners are generally less flexible than learning-based approaches, since they rely on explicitly defined assumptions about the environment, scale poorly to large or complex environments, and often fail to generalize across diverse conditions \citep{Popovic2024}. RL-based approaches, in contrast, can learn more general strategies from data, making them more adaptable to varying conditions and environments.

\subsection{Use in agricultural applications}
Weed detection was used as the use-case in this study, since weeds often occur in spatially distinct patches rather than being uniformly distributed across a field \citep{Garibay2001, Colbach2000}, but the approach is applicable to other agricultural tasks that involve locating non-uniformly distributed objects, such as disease detection, where infected plants often occur in spatially clustered patches rather than being spread uniformly across a field \citep{Campbell1985}. For applications where finding all objects is critical, a full coverage planner remains more suitable. However, for applications such as weed detection, where finding the majority of objects quickly is more important than exhaustive coverage, the simulation-trained RL path planner is efficient and shows potential for practical deployment on real drones.

\subsection{Limitations}
The real-world evaluation was limited in scale: experiment~1 used seven real-world datasets, each evaluated with four grid rotations (28 repetitions in total), and experiment~2 involved two flight trials. While these results already demonstrate that the simulation-trained policy transfers to real-world conditions, additional trials across a wider range of fields would help confirm how well performance generalizes to different field layouts, weed densities, and lighting conditions.

Furthermore, plastic plants were used as a proxy for real weeds in the real-world data, allowing repeatable experiments with known ground-truth locations. Real weeds exhibit more variety in shape, color, and growth stage than the plastic proxy, which the detection network would additionally need to handle. However, the detection network was already imperfect even on the plastic proxy, so the used proxy may still give a reasonably fair indication of the errors to be expected when detecting real weeds. Future work should nonetheless validate the approach on real weeds to confirm that detection and path planning performance generalizes to actual field conditions.

\subsection{Future work}
Increasing the reward for detecting an object during training would shift the recall-flight-path trade-off toward higher recall, but probably at the cost of a longer flight path. Similarly, improving the prior knowledge quality may also increase the recall of the RL agent, though again at the cost of a longer flight path. Improving the detection network would also help close the recall gap with the full coverage planner, although some gap would remain since the RL agent does not aim for full coverage.

The state- and action space of the RL agent was designed to work at a single altitude, corresponding to a fixed field of view. However, increasing the altitude of the drone increases the field of view of the camera and thereby increases efficiency, assuming that the recall of the detection network remains reasonable. The current detection network can already handle differences in altitude, as it was trained from images from multiple altitudes; however, it makes more mistakes at a higher altitude. By adapting the simulation environment, encoding detection confidence and altitude into the state space, and extending the action space with actions to increase and decrease the drone's altitude, the RL agent could potentially learn to control the altitude. For example, it could fly at a high altitude when the detection network is confident and lower its altitude when there are uncertain detections, thereby decreasing the needed flight time.

For practical deployment, the detection network and RL policy would typically run onboard the drone rather than on an external laptop as used in this study. In experiment~2, inference for both networks was executed on an external laptop on the ground, connected to the drone during flight, and the combined inference time was not measured here; this should be quantified in future work targeting fully onboard deployment. As an indication, the used detection network, YOLOv11-m, has an inference time of around 20ms on a NVIDIA Jetson Orin NX after quantization to INT8 \citep{Rey2025}. Since the used DQN is even smaller, running both networks onboard a drone should be feasible.

\section{Conclusion}
This paper evaluated a simulation-trained RL agent for adaptive drone path planning and demonstrated a proof-of-concept deployment on a real drone. In simulation, the RL agent achieved a 57\% shorter flight path compared to a full coverage planner, at the cost of a 13\% lower recall. On real-world data, the agent achieved a 38\% shorter flight path, at the cost of a 21\% lower recall. In the real-world drone flights, the agent found 73\% and 23\% of the weeds in trials~1 and~2, respectively.

The simulation-to-reality gap was mainly caused by differences in weed distributions and by errors made by the real detection network. In the real-world drone flight experiment, the drop in performance was mainly caused by the prior knowledge: when prior knowledge recall dropped to 0\%, meaning that the prior knowledge map provided no guidance, the agent found only 23\% of the weeds. Using detections from previous flights as prior knowledge could improve performance while removing the need for a dedicated high-altitude coverage flight.

For applications where finding all objects is critical, a full coverage planner remains more suitable. However, for applications such as weed and disease detection, where finding the majority of weeds quickly is more important than exhaustive coverage, the simulation-trained RL path planner shows potential for practical deployment on real drones. Further improvements to prior knowledge quality and flight path efficiency, along with validation across a wider range of real-world conditions, are needed before the approach can be reliably deployed in practice.

\section*{CRediT author statement}
\textbf{Rick van Essen:} Conceptualization, Methodology, Formal analysis, Software, Visualization, Writing - Original Draft. \textbf{Eldert van Henten:} Writing - Review \& Editing, Funding acquisition, Supervision. \textbf{Gert Kootstra:} Conceptualization, Methodology, Writing - Review \& Editing, Funding acquisition, Supervision.

\section*{Funding}
This research is part of the research program SYNERGIA (project number 17626), which is partly financed by the Dutch Research Council (NWO).

\section*{Data availability}
Four of the real-world datasets used in Experiment~1 are available at \url{https://doi.org/10.4121/bbe97051-07df-4934-b634-701d91a2075e}. The remaining three datasets are available upon request. The code to reproduce the results of this paper is available at \url{https://github.com/WUR-ABE/rl_drone_object_search_v2}. The code used to control DJI drones via MAVLink is available at \url{https://github.com/WUR-ABE/mavdrone}.

\section*{Declaration of generative AI and AI-assisted technologies in the manuscript preparation process}
Statement: During the preparation of this work the author(s) used Claude Sonnet 5 and Opus 4.6 in order to correct grammar and increase readability of the text. ChatGPT 4 and 5 were used to assist in coding. After using this tool/service, the authors reviewed and edited the content as needed and take full responsibility for the content of the published article and code.
\bibliographystyle{elsarticle-harv}
\bibliography{references}

\end{document}